%% file: main.tex
\PassOptionsToPackage{dvipsnames,svgnames,table}{xcolor}

\documentclass[]{meituan_xiaotuan}

\usepackage[T1]{fontenc}
\usepackage[utf8]{inputenc}   %
\usepackage{inconsolata}      %

\usepackage{graphicx}
\usepackage{float}
\usepackage{wrapfig}
\usepackage{needspace}
\usepackage{xcolor}
\usepackage{amsmath,amssymb}
\usepackage{bbm}

\usepackage{booktabs}
\usepackage{tabularx}
\usepackage{array}
\usepackage{multirow}
\usepackage{longtable}
\usepackage{makecell}
\usepackage{colortbl}

\usepackage{enumitem}
\usepackage[most]{tcolorbox}
\tcbuselibrary{breakable,skins}

\definecolor{takeawayFrame}{RGB}{165,209,216}
\definecolor{takeawayTitle}{RGB}{57,89,163}
\newtcolorbox{takeaway}[1][Takeaway]{%
  enhanced, breakable,
  colback=white, colframe=takeawayFrame, boxrule=0.8pt, arc=2mm,
  colbacktitle=white, coltitle=takeawayTitle,
  fonttitle=\sffamily\bfseries, title={#1}}

\usepackage[linesnumbered,ruled,vlined]{algorithm2e}

\usepackage{xspace}
\usepackage{url}
\usepackage{hyperref}
\hypersetup{
  colorlinks=true, linkcolor=mtblue, citecolor=mtblue, urlcolor=mtblue,
  pdftitle={ATLAS: Dual-Horizon Diagnostic Evaluation for Tool-Use Agents in Meituan Xiaotuan},
  pdfauthor={Wei Chen, Peilun Zhou, Zhaoyu Hu, Jiajun Chai, Derong Xu, Zhongni Hou, Yufei Zhang, Guojun Yin, Wei Lin, Zhi Zheng, Tong Xu},
  pdfsubject={ATLAS Technical Report},
  pdfkeywords={Agent Evaluation, Capability Diagnosis, Tool-Use Agents}
}

\newcommand{\Description}[1]{}

\newcommand{\gradname}{ATLAS}

\title{\textbf{\gradname: Dual-Horizon Diagnostic Evaluation for Tool-Use Agents in Meituan Xiaotuan}}

\author{\textbf{University of Science and Technology of China}}
\author{  \textbf{Meituan}}
\affiliation{See \hyperref[sec:contribution]{Contributions} section for the full author list.}

\abstract{
Large language model (LLM)-based assistants on web-scale service platforms increasingly address everyday local-service needs through multi-step planning and tool use. Under dynamic business conditions and evolving user needs, outcome-level evaluation provides limited evidence of where execution falls short, which quality or compliance requirement is unmet, or how service loses alignment across continued interaction, making targeted iteration difficult. We present ATLAS, a dual-horizon diagnostic evaluation framework for industrial tool-use agents. ATLAS organizes diagnosis around two complementary horizons: the complete agent lifecycle triggered by a user request, and continued user interaction spanning multiple such lifecycles. For a single lifecycle, within-turn evaluation uses trajectory-wise signals across three execution dimensions (thinking and reflection, tool and skill execution, and response generation) to identify where behavior falls short and which capability concerns are unmet. Across lifecycles, across-turn evaluation uses user-wise signals to diagnose failures in carrying forward or updating user intent, constraints, and corrections. These fine-grained diagnostic signals ground diagnosis in explicit evidence scopes and decision boundaries, using rule-based checks and LLM judgments. To support reliable diagnosis at production scale, LLM judges are calibrated on dynamically maintained references from business logs, and selected judges are distilled into compact diagnostic models. The resulting diagnostic profiles help developers attribute failures across execution stages, prioritize improvement targets, and reassess subsequent interventions. Applied in Meituan Xiaotuan, a million-scale DAU local-services assistant, ATLAS achieves high diagnostic fidelity across calibrated judges and distilled models. Furthermore, blinded human auditing and a 14-day online A/B test (nearly two million users per arm) confirm that the ATLAS-guided agent improves recommendation relevance, factual reliability, user engagement (+2.84\% answer read-through rate), and order volume (+6.13\% paid orders), leading to full production deployment.
}

\correspondence{\email{chenweicw@mail.ustc.edu.cn, zhoupeilun02@meituan.com}}
\checkdata[Project Page]{\url{https://atlas-eval.github.io/}}

\begin{document}
\maketitle

\input{sections/1-introduction}
\input{sections/2-related-work}
\input{sections/3-method}
\input{sections/4-experiments}
\input{sections/5-conclusion}

\clearpage
\noindent\begin{minipage}{\linewidth}
\small
\setlength{\parindent}{0pt}
\setlength{\parskip}{2pt}
\phantomsection
\section*{Contributions}
\label{sec:contribution}
\addcontentsline{toc}{section}{Contributions}

\textbf{Authors}
Wei Chen$^{1,*}$,
Peilun Zhou$^{2,*,\ddagger}$,
Zhaoyu Hu$^{2}$,
Jiajun Chai$^{2}$,
Derong Xu$^{1}$,
Zhongni Hou$^{2}$,
Yufei Zhang$^{2}$,
Guojun Yin$^{2,\dagger}$,
Wei Lin$^{2,\dagger}$,
Zhi Zheng$^{1,\dagger}$,
Tong Xu$^{1,\dagger}$

\textbf{Affiliations}
$^1$\,State Key Laboratory of Cognitive Intelligence, University of Science and Technology of China;
$^2$\,Meituan

$^{*}$ Equal Contribution; $^{\ddagger}$ Project Leader; $^{\dagger}$ Corresponding Author
\end{minipage}\par

\clearpage
\bibliographystyle{plainnat}
\bibliography{refs}

\clearpage
\beginappendix
\input{sections/appendix}

\end{document}

%% file: sections/1-introduction.tex
\section{Introduction}
\begin{figure}[!t]
    \centering
    \includegraphics[width=0.75\linewidth]{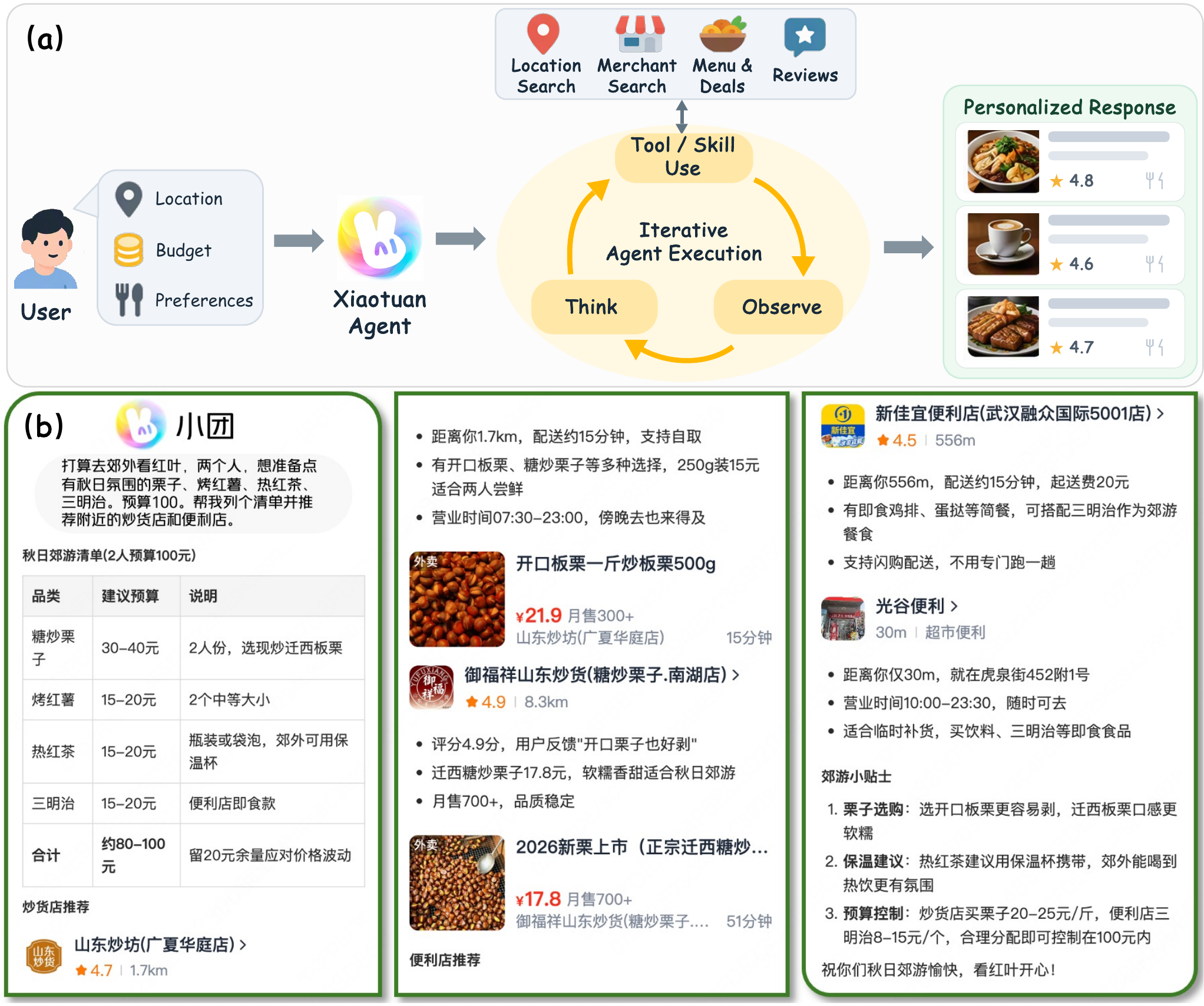}
    \caption{An example interaction in Meituan Xiaotuan: (a) an illustrative multi-step agentic workflow, and (b) the user-facing service response.}
    \label{fig:xiaotuan_case}
\end{figure}

Large language model (LLM) agents are increasingly deployed across web-scale service platforms, shifting online information access from static item retrieval toward interactive, tool-augmented assistance~\cite{yao2023react,schick2023toolformer,xu2026aiagentsys,he2025vitabench,wu2026aligning,liao2025reclm,deldjoo2025holistic}.
Unlike standalone text generators, production service agents operate as compound systems that interleave multi-step reasoning, external tool and skill invocation, and live supply synthesis.
When service quality falls short in deployment, defects may stem from workflow orchestration, tool interfaces, scenario data coverage, or policy behavior, making blind trial-and-error costly under finite engineering and training resources.
Effective iteration therefore requires diagnostic evaluation as an operational compass rather than a static leaderboard~\cite{prm800k,guo2026agenteval}: it must answer \textbf{what fails, what should be prioritized, and whether an intervention worked}.

Existing evaluation paradigms only partially support this diagnostic need.
Standard agent benchmarks primarily measure terminal task completion or final state matching~\cite{yao2024taubench,barres2025tau2bench,jimenez2024swebench,yu2026wildtoolbench,zhu2026toolmaze}, offering little visibility into intermediate execution.
Recent process-oriented evaluations step inside the trajectory to assess intermediate reasoning and action steps~\cite{prm800k,guo2026agenteval,men2025agentrewardbench}, while rubric-based approaches decompose open-ended response quality into fine-grained checklist criteria~\cite{gunjal2025rubrics,chen2026holistic,mehta2026complexconstraints,xu2026rtt}.
However, existing process checks are largely designed for static environments, whereas rubric evaluations typically focus on isolated final responses, leaving a gap for production agents across dynamic tool workflows and continued interactions.

This gap is particularly evident in Meituan Xiaotuan, an AI assistant for local services with a million-scale daily active user base.\footnote{Across a platform of hundreds of millions of active users, Xiaotuan is backed by 700 million nationwide merchant-information verifications, 1.3 billion real-review recalibrations, and more than 3.4 million merchant uses of its AI-assistant functions.}
As illustrated in Figure~\ref{fig:xiaotuan_case}, behind the user-facing response (b), fulfilling a request triggers an iterative execution workflow (a) in which the agent reasons over user needs, invokes production tools and skills to retrieve live business information, and synthesizes a grounded service response (see Appendix~\ref{app:xiaotuan_platform}).
Evaluating and iterating such an agent in production raises three core challenges:
\textbf{1) Cascading errors across multi-step execution.} Within a single request lifecycle, an early deviation in intent understanding, query rewriting, or tool selection propagates through subsequent steps into an off-target final response~\cite{gurram2026agentpropbench}; outcome-only evaluation cannot tell whether the failure originated in final answer synthesis or upstream execution.
\textbf{2) Multifaceted quality requirements under dynamic supply.} Because open-ended requests admit multiple valid paths while merchant inventory, pricing, and availability change continuously~\cite{song2026mobilitybench}, quality cannot be judged against a fixed reference answer. Instead, a useful recommendation must simultaneously satisfy diverse business and compliance requirements, such as user constraints, retrieved supply facts, deal timeliness, logical consistency, and safety rules~\cite{liu2026openrubrics}, where violating any single aspect can leave a response inaccurate, outdated, or unactionable.
\textbf{3) Context drift over continued interaction.} Beyond a single request, service unfolds across multi-turn sessions where prior exchanges establish referents, preferences, constraints, and explicit corrections. Even if each lifecycle appears locally plausible, failing to carry forward or update this evolving state causes service to drift from the user's actual goal.

To address these challenges, we propose \textbf{ATLAS},\footnote{Named after a cartographic atlas, ATLAS maps agent behavior and capability deficiencies across complementary execution and interaction horizons.} a dual-horizon diagnostic evaluation framework that organizes capability analysis and downstream improvement around two complementary scopes.
\textbf{Within-Turn Evaluation} uses trajectory-wise diagnostic signals to examine the complete agent lifecycle triggered by a user request, crossing three \textit{execution dimensions} (\textit{Thinking \& Reflection}, \textit{Tool \& Skill Execution}, and \textit{Response Generation}) with five recurring \textit{capability concerns} (\textit{Relevance}, \textit{Factuality}, \textit{Timeliness}, \textit{Reliability}, and \textit{Intent \& Planning}) alongside an independent \textit{Norms \& Compliance} guardrail to jointly pinpoint which execution stage fails and which quality or compliance requirement is violated.
\textbf{Across-Turn Evaluation} complements this trajectory-wise matrix with user-wise diagnostic signals spanning multiple lifecycles, conditioning on prior turn context and the current trajectory to diagnose whether referents, user intent, constraints, and explicit corrections are properly carried forward and updated.

To make this dual-horizon structure actionable and maintainable in production, ATLAS instantiates its taxonomy into fine-grained rule- and LLM-based signals grounded in explicit trajectory evidence and decision boundaries.
LLM-based signals are calibrated on rolling, dynamically maintained references from business logs so that decision boundaries co-evolve with changing tools and failure modes, while high-frequency judges are distilled into grouped compact models for scalable evaluation.
In a human-in-the-loop iteration workflow, summarizing signal outcomes across traffic yields a multi-dimensional diagnostic profile that helps developers trace surface failures to upstream bottlenecks, select targeted interventions (e.g., workflow refinement, data curation, or policy optimization), and reassess subsequent iterations.
In Xiaotuan, offline evaluations confirm that ATLAS achieves high diagnostic fidelity across calibrated judges and distilled models, while blinded human auditing and a 14-day online A/B test (nearly two million users per arm) verify that the ATLAS-guided agent significantly improves recommendation quality, user engagement, and paid orders, leading to its full production deployment.

In summary, our contributions are threefold:
\begin{itemize}[leftmargin=*]
    \item \textbf{Dual-Horizon Diagnostic Evaluation:} We introduce ATLAS, a diagnostic framework that combines trajectory-wise within-turn signals across execution dimensions and capability concerns with user-wise across-turn signals over multiple lifecycles.

    \item \textbf{Scalable Calibration and Diagnosis-Guided Iteration:} ATLAS operationalizes its taxonomy through rule- and LLM-based signals calibrated on rolling production references, distills grouped compact models for scalable evaluation, and guides human-in-the-loop downstream iteration.

    \item \textbf{Industrial Validation and Deployment:} In Meituan Xiaotuan, ATLAS achieves high diagnostic fidelity and drives agent gains in recommendation quality, user engagement, and paid orders across blinded human auditing and a 14-day online A/B test (nearly two million users per arm), with the improved agent deployed to all production traffic.
\end{itemize}

%% file: sections/2-related-work.tex
\section{Related Work}
\subsection{Tool-Use Agent Evaluation}
Tool-use agents operate through sequences of decisions that connect a user request with changing external evidence, tool outputs, and a final service response.
Accordingly, evaluation has developed along complementary directions.
General-purpose benchmarks measure end-to-end performance through task completion, terminal-state matching, or patch validation, providing common and reproducible measures of overall agent success~\cite{yao2024taubench,barres2025tau2bench,jimenez2024swebench,yu2026wildtoolbench}.
At the same time, process-oriented studies retain intermediate execution evidence to examine how an outcome is produced.
For instance, AgentEval models workflows as dependency-aware DAGs for failure analysis~\cite{guo2026agenteval}, and Agent-RewardBench evaluates reward models across perception, planning, and safety~\cite{men2025agentrewardbench}.
This growing emphasis on execution evidence is particularly consequential for open-ended, multi-step agents, where a final outcome alone may not expose the decisions and conditions that shaped it.
Signal design is therefore critical: LLM evaluators vary substantially across agent capabilities~\cite{men2025agentrewardbench}, and coarse judgments often miss defects in deployed multi-turn transactional agents~\cite{zhang2026catching}.

Another line of work grounds agent evaluation in real interaction logs and domain-specific environments, including mobility route planning~\cite{song2026mobilitybench}, product-facing deep research~\cite{minddr2026}, recommendation reward modeling~\cite{zeng2026recrmbench}, and multi-session health assistance~\cite{zhang2026rubricstree}.
ATLAS builds on this landscape by uniting two complementary scopes in one diagnostic framework: within-turn evaluation localizes deficiencies across the trajectory of a single lifecycle, while across-turn evaluation diagnoses whether user context and corrections are maintained across lifecycles.

\subsection{Diagnostic Signals for Agent Iteration}
The signals used to optimize agents have likewise evolved from outcome-level rewards toward richer forms of supervision.
Terminal task success and auxiliary objectives remain common sources of feedback for tool-use reinforcement learning~\cite{qian2026toolrl}.
Process supervision instead assigns credit to intermediate reasoning steps or actions~\cite{wang2023mathshepherd,prm800k}, and tool-use RL further constructs local feedback around invocation and execution behavior~\cite{qian2026toolrl}.
Rubric-based approaches extend structured criteria to tasks without a single verifiable outcome~\cite{gunjal2025rubrics,chen2026holistic,mehta2026complexconstraints}, while work such as Rubrics to Tokens connects response-level rubrics with finer-grained training signals~\cite{xu2026rtt}.
In industrial and domain-specific applications, multi-dimensional rubric feedback has been used to guide safety alignment~\cite{loye2026rubas}, web search~\cite{wu2026aligning}, sponsored-search ad generation~\cite{wei2026interactor}, e-commerce dialogue~\cite{li2026more}, and recommendation agents~\cite{yi2025recgptv2}.

While prior studies primarily use rubric or process signals as direct rewards for a single training stage, ATLAS uses calibrated signals across the full iteration loop: diagnostic profiles first locate execution or cross-turn bottlenecks to guide workflow refinement, data curation, or policy optimization, and selected signals and compliance guardrails can then be reused as gated training feedback before offline and online reassessment in production.

%% file: sections/3-method.tex
\setcounter{topnumber}{1}
\section{Methodology}
\label{sec:methodology}
\begin{figure}[t]
    \centering
    \includegraphics[width=0.94\textwidth]{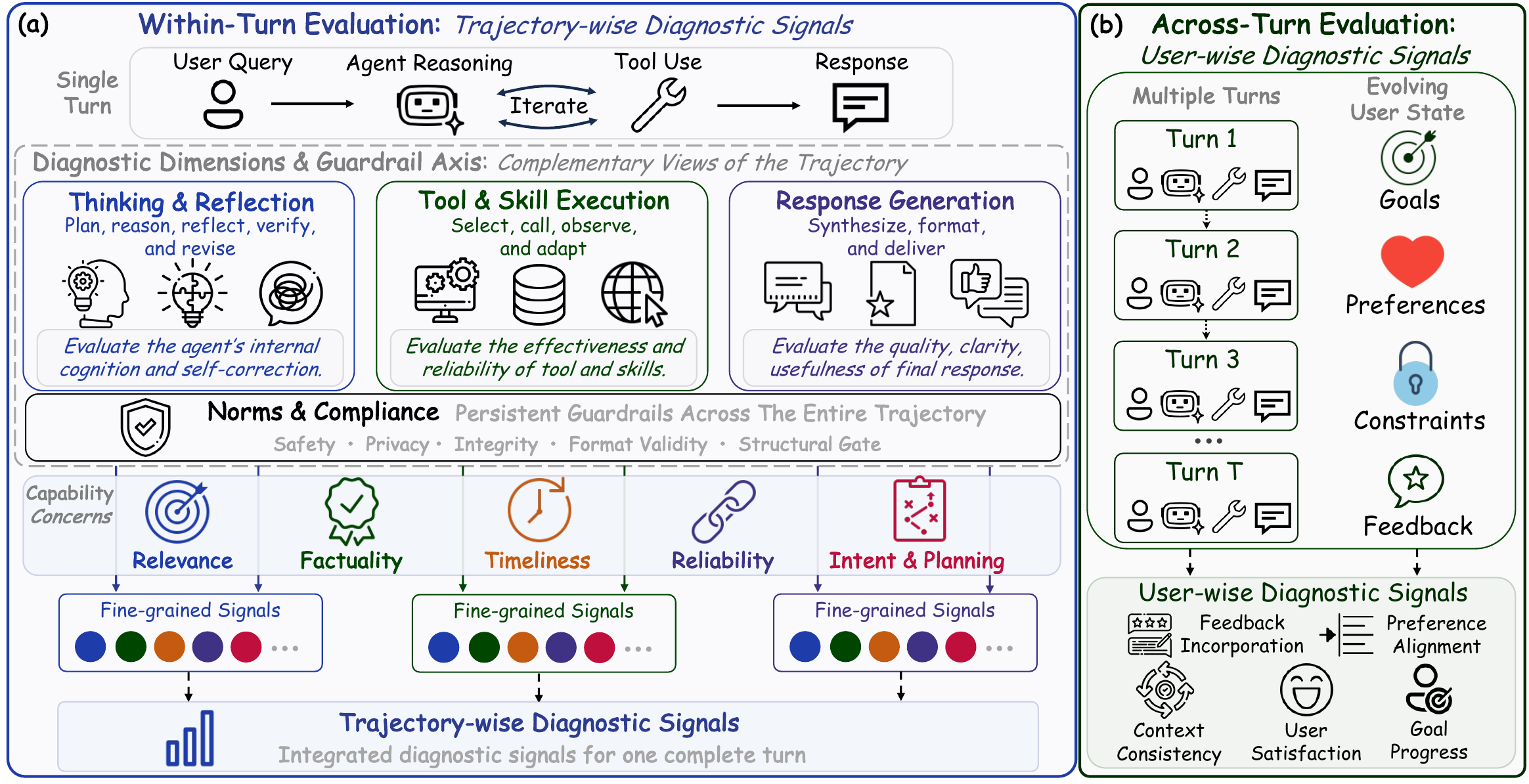}
    \caption{Overview of ATLAS's dual-horizon diagnostic evaluation: (a) within-turn evaluation applies trajectory-wise signals to a single agent lifecycle, and (b) across-turn evaluation applies user-wise signals across multiple lifecycles.}
    \label{fig:atlas_dual_horizon}
\end{figure}

To connect fine-grained diagnosis with production iteration, ATLAS comprises five components: the dual-horizon diagnostic structure (\S\ref{sec:diagnostic_structure}), signal construction (\S\ref{sec:signal_construction}), rolling calibration (\S\ref{sec:signal_calibration}), model distillation (\S\ref{sec:model_distillation}), and diagnosis-guided iteration (\S\ref{sec:agent_iteration}).

\subsection{Dual-Horizon Diagnostic Structure}
\label{sec:diagnostic_structure}
\begin{figure}[t]
    \centering

    \includegraphics[width=0.95\textwidth]{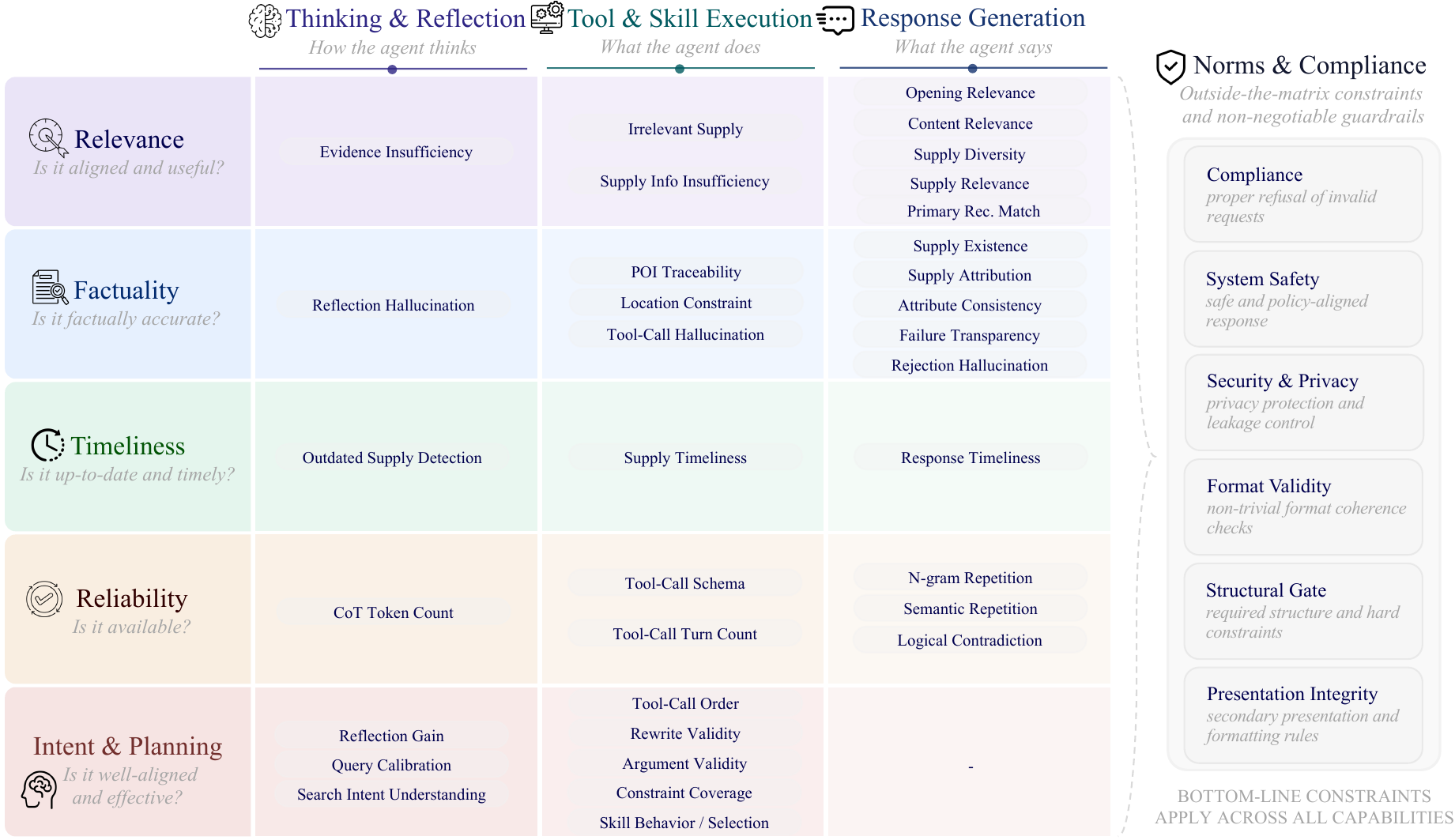}
    \caption{Within-turn diagnostic matrix in ATLAS, organizing trajectory-wise signals by execution dimension and capability concern alongside the independent Norms \& Compliance guardrail.}
    \label{fig:atlas_taxonomy}

\end{figure}

ATLAS is built on the observation that industrial agent service unfolds over two connected temporal scopes (Figure~\ref{fig:atlas_dual_horizon}).
\emph{Within-turn evaluation} examines the complete agent lifecycle (execution trajectory) initiated by a user request, during which the agent iteratively reasons, invokes tools or skills, and adapts to tool feedback before delivering a final response.
Preserving this process evidence makes it possible to determine where a deficiency arises and how it falls short of task requirements.
Within-turn evaluation organizes trajectory-wise signals in the $3\times 5$ diagnostic matrix of Figure~\ref{fig:atlas_taxonomy}, crossing the three \emph{execution dimensions} (\emph{Thinking \& Reflection}, \emph{Tool \& Skill Execution}, and \emph{Response Generation}) with five \emph{capability concerns} (\emph{Relevance}, \emph{Factuality}, \emph{Timeliness}, \emph{Reliability}, and \emph{Intent \& Planning}), alongside an independent \emph{Norms \& Compliance} guardrail for non-negotiable safety, privacy, and structural constraints.
These dimensions and concerns are instantiated by fine-grained diagnostic signals with explicit evidence scopes and decision boundaries (Appendix Tables~\ref{tab:reward_catalog_generation}--\ref{tab:reward_catalog_userwise}), forming an extensible diagnostic backbone that scales as product capabilities evolve.

\emph{Across-turn evaluation} complements this trajectory-wise matrix by examining a sequence of lifecycles within an ongoing interaction. Conditioned on prior turn context and the current lifecycle trajectory (Figure~\ref{fig:atlas_dual_horizon}(b)), ATLAS instantiates cross-turn continuity objectives into six user-wise signals (Appendix Table~\ref{tab:reward_catalog_userwise}) that diagnose reference resolution, reasoning and tool alignment, previous-round correction, and cross-round contradiction or repetition.

\FloatBarrier
\Needspace{22\baselineskip}
\begin{wrapfigure}{r}{0.50\textwidth}
    \centering
    \captionsetup{font=footnotesize}
    \includegraphics[width=\linewidth]{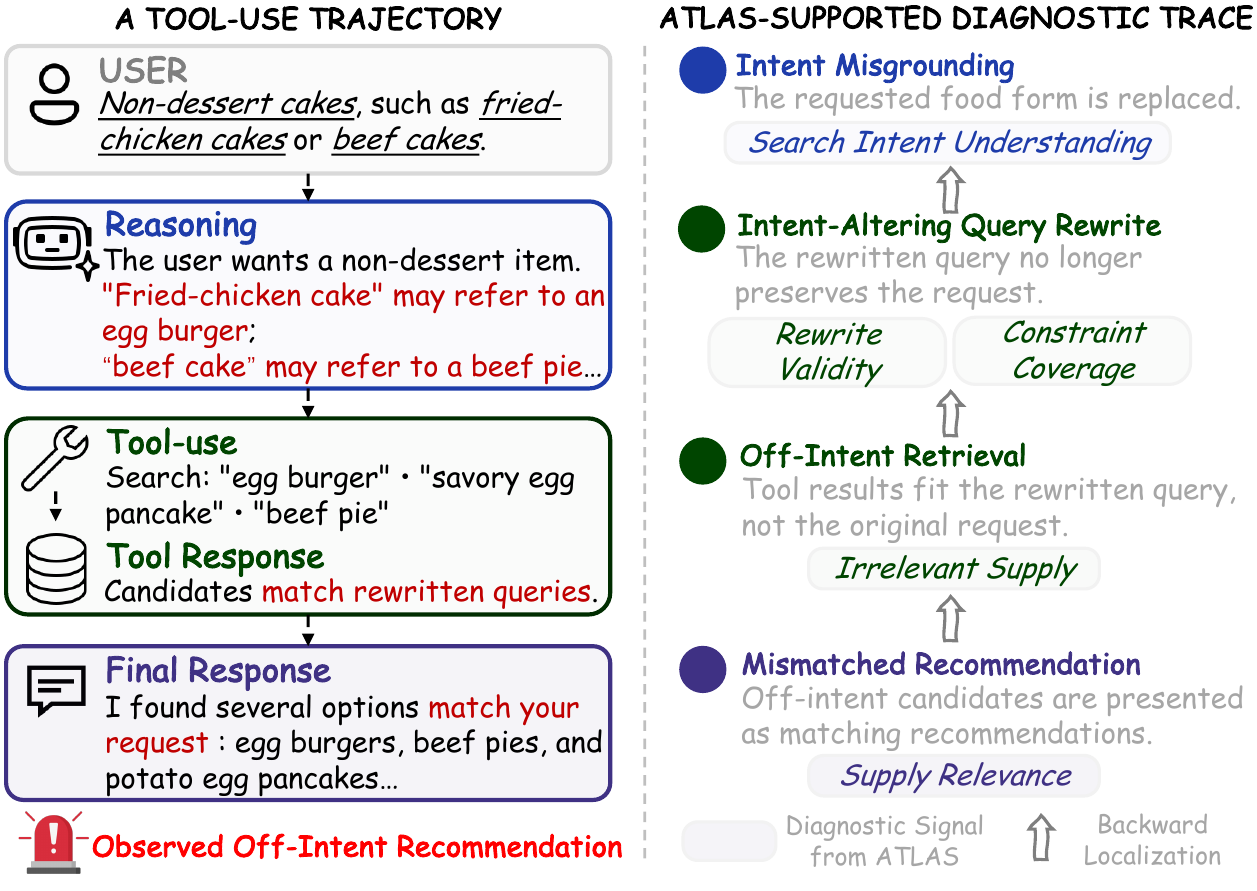}
    \caption{Cross-dimension fault attribution with ATLAS, tracing an off-intent recommendation back to intermediate query-rewrite and intent-understanding failures.}
    \label{fig:atlas_diagnostic_evidence}
\end{wrapfigure}

Crucially, evaluating these dimensions jointly separates downstream response symptoms from upstream execution errors. As shown in Figure~\ref{fig:atlas_diagnostic_evidence}, when a response recommends off-intent items (flagged by \emph{Supply Relevance} in \emph{Response Generation}), outcome-only evaluation cannot tell whether the failure arose in final synthesis or earlier execution. Tracing ATLAS signals along the trajectory shows that the response faithfully summarized retrieved items, while an upstream error occurred in \emph{Search Intent Understanding} (\emph{Thinking \& Reflection}) and \emph{Rewrite Validity} (\emph{Tool \& Skill Execution}), which replaced the requested food category. Aggregating these traces across traffic surfaces systematic bottlenecks.

\ifnum\value{WF@wrappedlines}>1
    \vspace{\dimexpr\value{WF@wrappedlines}\baselineskip-\baselineskip\relax}
\fi
\WFclear
\subsection{Diagnostic Signal Construction}
\label{sec:signal_construction}
\begin{wrapfigure}{r}{0.46\textwidth}
    \centering
    \captionsetup{font=footnotesize}
    \includegraphics[width=\linewidth]{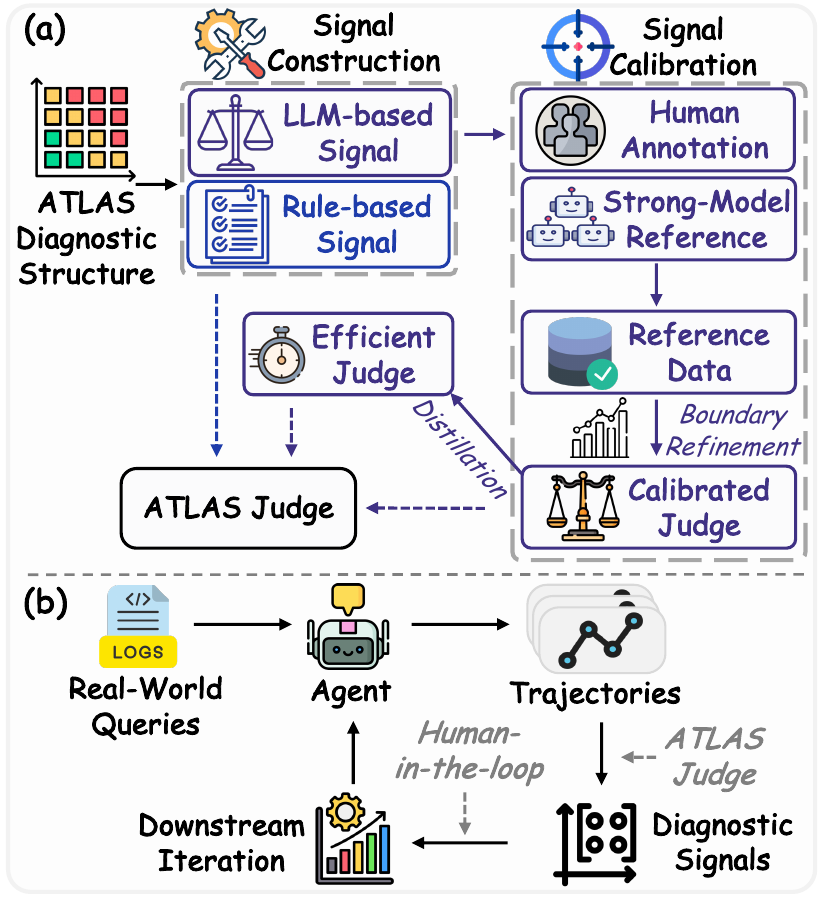}
    \caption{Diagnostic signal construction and iteration workflow in ATLAS: (a) Dual-Horizon Signal Construction via rule checks, rolling judge calibration, and model distillation; (b) Diagnosis-Guided Agent Iteration via targeted interventions.}
    \label{fig:atlas_diagnostic_pipeline}
\end{wrapfigure}

ATLAS instantiates the diagnostic structure in Section~\ref{sec:diagnostic_structure} as executable signals (Figure~\ref{fig:atlas_diagnostic_pipeline}(a)).
Each signal specifies a localized diagnostic target, the admissible evidence scope, an explicit decision boundary, and structured output semantics, implemented via either programmatic rules or calibrated LLM judges.

\Needspace{26\baselineskip}
\noindent\textbf{Signal implementation by evidence type.}
Rule-based diagnostic signals are used when the relevant boundary can be verified programmatically, such as for format constraints, structural validity, explicit enumerations, fixed-schema legality, or deterministic counts.
LLM-based diagnostic signals are used when assessment requires open semantic reasoning over evidence that may jointly involve the user query, conversational context, tool outputs, intermediate reasoning, or the final response.
For such signals, the LLM judge turns the diagnostic specification into a review task with explicit admissible evidence, decision boundaries, and structured outputs.
Each judge prompt is structured around five components: (i) \emph{role and task definition}, (ii) \emph{input field specification} binding the admissible evidence and target objects, (iii) \emph{decision criteria with explicit exemption boundaries} (supplemented by few-shot near-miss examples when inputs are heterogeneous), (iv) a \emph{stepwise review procedure} covering evidence extraction, local comparison, and boundary checking, and (v) a \emph{structured output format} that standardizes local verdicts, counts, and rationales.

\noindent\textbf{Diagnostic score formulation.}
Signals produce binary pass/fail verdicts by default, while targets that admit quantitative criteria return local judgments, counts, or rankings for post-processing.
Formally, for an input context $x$ and trajectory $\tau$ under signal $m$, the evaluator identifies a set of locally assessable objects $A_m(x,\tau)=\{a_{m,1},\ldots,a_{m,K}\}$ (such as claims, candidate supply items, typo positions, or reflection steps).
The evaluator then produces local judgments or sufficient statistics, and the final diagnostic score is defined as
\begin{equation}
    s_m(x,\tau)=\phi_m\big(z_{m,1}(x,\tau),\ldots,z_{m,K}(x,\tau)\big),
    \label{eq:signal_scalarization}
\end{equation}
where $z_{m,k}(x,\tau)$ denotes the local verdict or derived statistic associated with object $a_{m,k}$, and $\phi_m$ is a metric-specific normalization function that maps local judgments to a bounded scale $[0,1]$ with higher values indicating better compliance (e.g., computing the proportion of valid items rather than summing raw counts over $K$, and excluding trajectories where $A_m(x,\tau)$ is empty).
This construction yields interpretable scores for signals such as repetition rate, stale-item ratio, correction rate, or ranking quality, because each aggregate remains grounded in local assessments rather than a holistic scalar judgment.

\subsection{Signal Calibration and Usability}
\label{sec:signal_calibration}
While rule-based signals are determined by their programmatic checks, each LLM-based signal requires calibration to ensure that its judge interface reproduces the intended decision boundary on production-aligned reference data.

\noindent\textbf{Reference construction and stratified sampling.}
For each LLM-based diagnostic signal, ATLAS dynamically maintains a reference set of approximately $1,000$ examples sampled from production logs over a rolling six-month window.
To ensure broad coverage across real-world operating conditions, we stratify sampling along three dimensions: (i) \emph{business scenarios} (e.g., food delivery, in-store dining, instant retail, and healthcare), (ii) \emph{request complexity} (explicit single-constraint, multi-constraint, and ambiguous or underspecified queries), and (iii) \emph{interaction requirements} (follow-up questions, clarifications, cross-turn coreference, and skill invocations).
Within each signal, binary labels are balanced at approximately $1{:}1$, while continuous signals are sampled evenly across score bins (e.g., intervals of $0.2$ on $[0,1]$), supplemented by controlled query- and trajectory-level perturbations for low-frequency safety or compliance boundaries.
Reference labeling follows the evidence structure of each target.
For signals in \emph{Response Generation}, five senior annotators follow a unified protocol, each sample receives at least three independent labels, and disputed cases are adjudicated by domain experts.
For signals involving long tool-use trajectories, extensive tool outputs, or complex evidence chains, we retain unanimous verdicts across three strong teacher models and conduct stratified human spot-checking (requiring a batch verification accuracy above $95\%$) to verify evidence grounding and filter out cases where the teachers share common blind spots.

\noindent\textbf{Usability criterion, boundary refinement, and rolling maintenance.}
Each signal's reference set is partitioned by original conversation into a $1{:}1$ stratified split: a calibration split $\mathcal{D}^{\mathrm{cal}}_m$ for iterative prompt refinement and a held-out test split $\mathcal{D}^{\mathrm{test}}_m$ with matched scenario and label distributions for reporting diagnostic fidelity.
We use F1 as the primary criterion (with continuous scores discretized into pass/fail labels) and admit a judge interface $j_m$ once its calibration performance satisfies
\begin{equation}
    \mathrm{F1}\big(j_m, \mathcal{D}^{\mathrm{cal}}_m\big) \geq \tau_m,
    \label{eq:judge_usability}
\end{equation}
where $\tau_m$ is set in a high-confidence range, typically $0.90$.
When F1 on $\mathcal{D}^{\mathrm{cal}}_m$ falls below $\tau_m$, we trace error cases to resolve boundary ambiguity, missing evidence specifications, or insufficient exemption rules, and freeze the refined interface before evaluating it on $\mathcal{D}^{\mathrm{test}}_m$.
In continuous production use, as the six-month log window advances, newly audited failure cases replace older examples and trigger boundary updates when needed, allowing ATLAS to adapt to evolving user tasks, tool schemas, and business rules.

\subsection{Diagnostic Model Distillation}
\label{sec:model_distillation}
Although calibrated LLM judges provide high-capacity semantic assessment, invoking them repeatedly at scale incurs substantial latency and cost. For selected LLM-based signals, ATLAS therefore distills compact diagnostic models that reproduce the calibrated decision behavior on the same evidence inputs and structured outputs at a fraction of the inference cost.

\noindent\textbf{Grouping strategy.}
We adopt group-wise distillation rather than fully independent or fully shared training.
Independent models preserve individual decision boundaries but fragment data, duplicate parameters, and increase deployment overhead, whereas a fully shared model can mix signals with incompatible evidence forms and output targets.
Distillation groups are therefore organized jointly by \emph{execution dimension}, \emph{capability concern}, and \emph{input--output form}: the first two preserve a coherent diagnostic perspective, while the third aligns the prediction interface.

\noindent\textbf{Data and supervision.}
Distillation training trajectories are sampled from production logs across diverse scenarios and task forms.
On a subset of training trajectories, we apply controlled \emph{response-level perturbation}: the final response is lightly rewritten while preserving the core task semantics and content, introducing variation in wording, style, and formatting.
This augmentation encourages the diagnostic model to learn stable decision behavior rather than overfit to a particular surface template.
Multiple strong teacher models then independently assess the resulting trajectories together with the original request and execution evidence, and majority voting produces the distillation supervision.
Distillation training trajectories are separated by original conversation from the held-out evaluation splits in $\mathcal{D}_{\text{Judge}}$, ensuring that student models are evaluated against independent reference instances.

\noindent\textbf{Filtering and objective.}
Before training, we apply \emph{format consistency filtering} to remove structurally abnormal samples, including failed JSON parsing, unclosed \texttt{<think>} tags, repetitive responses, and abnormally short reasoning traces.
We then apply \emph{confidence-based filtering}: when the base model's prediction already matches the target label with log-probability above a predefined threshold, the sample is treated as low-gain and removed from distillation.
Within each group, the diagnostic model is trained to map the corresponding evidence input to the teacher-derived decision output.
When a \texttt{<think>} field is present, it receives a lower loss weight, e.g., $0.1$, to reduce overfitting to teacher-specific reasoning style and place greater emphasis on the final verdict and key structured fields.
The resulting efficient diagnostic models support low-latency, lower-cost evaluation for large-scale log monitoring, regression testing, and iterative model optimization.

\subsection{Diagnosis-Guided Agent Iteration}
\label{sec:agent_iteration}
As shown in Figure~\ref{fig:atlas_diagnostic_pipeline}(b), aggregating signal outcomes across production or replay traffic yields a structured diagnostic profile that supports a human-in-the-loop iteration workflow.
The profile enables developers to pinpoint capability bottlenecks and select matching interventions: 
refining the agent harness, execution pipeline, tool schemas, or skill-routing rules when errors stem from workflow orchestration or interface constraints; curating targeted fine-tuning data or rebalancing training mixtures when deficiencies concentrate in specific scenarios or constraint types; or applying policy optimization when improvements require coordinating multi-step reasoning, tool grounding, and response generation.

In a given iteration cycle, once localized harness or data fixes are in place, the remaining bottlenecks often lie in jointly coordinating multi-step reasoning, tool-evidence grounding, and response generation under dynamic supply.
To address such cross-dimension trade-offs, as in the Xiaotuan iteration studied in this paper, ATLAS supports policy optimization by reusing selected calibrated diagnostic signals $\mathcal{S}_{\mathrm{train}}$ as multi-dimensional feedback while enforcing \emph{Norms \& Compliance} as a hard guardrail gate.
Formally, for an input query with context $x$ and a generated trajectory $\tau$, the gated training return is defined as
\begin{equation}
    R(x,\tau)=G_{\mathrm{norm}}(x,\tau)\cdot\sum_{m\in\mathcal{S}_{\mathrm{train}}}\lambda_m\, s_m(x,\tau),
    \label{eq:training_reward}
\end{equation}
where $G_{\mathrm{norm}}(x,\tau)\in\{0,1\}$ is a training-time gate ($1$ only if $\tau$ satisfies all \emph{Norms \& Compliance} signals, and $0$ otherwise), and $\lambda_m$ weights each applicable capability signal $s_m(x,\tau)$ according to the diagnosed iteration priorities.
The specific signal weights and Group Relative Policy Optimization (GRPO)~\cite{shao2024deepseekmath} training setup used in our Xiaotuan iteration are detailed in Appendix~\ref{app:impl_details}.
Following the intervention, the updated agent is reassessed through diagnostic profiles, independent human auditing, and online A/B testing to verify whether targeted deficiencies are resolved in production.
\FloatBarrier
\setcounter{topnumber}{3}

%% file: sections/4-experiments.tex
\section{Experiments}
We evaluate ATLAS along four research questions: \textbf{RQ1 (Diagnostic Fidelity)}, how accurately calibrated LLM judge interfaces enforce signal-specific decision boundaries; \textbf{RQ2 (Scalable Distillation)}, whether grouped compact models retain calibrated judgment fidelity at lower cost; \textbf{RQ3 (Human-Audited Quality)}, whether ATLAS-guided iteration improves recommendation quality under blinded human review; and \textbf{RQ4 (Online Impact)}, whether these gains transfer to user engagement and transactions in live traffic.

\subsection{Experimental Setup}
\input{tables/table_reward_judge_main}

\noindent\textbf{Datasets and evaluation environment.}
We construct all datasets from de-identified Meituan Xiaotuan production logs over a six-month window using the stratified sampling protocol in Section~\ref{sec:signal_calibration} (spanning business scenarios, request complexity, and multi-turn interaction types; see Appendix~\ref{app:xiaotuan_platform}) and separate them strictly by original conversation.
The \textbf{Diagnostic Signal Benchmark} ($\mathcal{D}_{\text{Judge}}$) covers the 41 LLM-based signals with approximately $1,000$ logged query--trajectory instances per signal (recording dialogue context, intermediate reasoning, tool calls and supply payloads, and final responses), partitioned $1{:}1$ into a calibration split $\mathcal{D}^{\mathrm{cal}}_m$ and a held-out test split $\mathcal{D}^{\mathrm{test}}_m$.
For RQ2, compact diagnostic models are trained on conversation-disjoint logged trajectories and evaluated on $\mathcal{D}^{\mathrm{test}}_m$.
For the downstream iteration study in RQ3--RQ4, policy optimization ($\mathcal{D}_{\text{RL}}$, ${\sim}10,000$ requests) and the $1,000$-query paired human audit sample user queries with their prior dialogue context, from which tested agents generate fresh tool-use trajectories.

\noindent\textbf{LLM judge interfaces.}
To isolate the role of explicit evidence binding, exemption handling, and stepwise review, we compare four interfaces constructed under the same DeepSeek-V4-Pro backbone and evaluated on $\mathcal{D}^{\mathrm{test}}_m$: \textbf{Direct Judge} retains only the task definition, input description, and output schema; \textbf{Static Rubric} adds an automatically generated fixed rubric; \textbf{Curated Rubric} compresses the calibrated signal-specific criteria and key exemptions into a compact rubric with a simplified review procedure; and \textbf{ATLAS Judge} uses the full calibrated specification (\S\ref{sec:signal_calibration}; Appendix~\ref{app:judge_interface_configurations}).

\noindent\textbf{Efficient diagnostic models.}
Following Section~\ref{sec:model_distillation}, we distill compact models for selected LLM-based diagnostic signals onto Qwen3.5-9B.
We compare the resulting efficient diagnostic models with the untuned Qwen3.5-9B backbone~\cite{qwen35blog}, Qwen3.6-27B~\cite{qwen3.6-27b}, Qwen3.6-35B-A3B~\cite{qwen36_35b_a3b}, and DeepSeek-V4-Pro~\cite{xu2026deepseek}.

\noindent\textbf{Metrics.}
For LLM judge-interface (RQ1) and efficient diagnostic model (RQ2) experiments, we report \textbf{F1} on the held-out splits $\mathcal{D}^{\mathrm{test}}_m$ (with group summaries reporting unweighted means across signals).
For the independent human audit (RQ3), we report \textbf{response--supply relevance}, \textbf{P0 hallucination rate}, and \textbf{ID hallucination rate} based on binary $0/1$ human judgments.
For the 14-day online A/B test (RQ4), we report \textbf{AI answer read-through rate}, \textbf{effective QV}, \textbf{merchant-card exposure QV}, \textbf{paid order count}, \textbf{paid GTV}, and \textbf{5-second session exit rate} (defined in Appendix~\ref{app:metrics}).

\noindent\textbf{Implementation details.}
All experiments run on NVIDIA H20 GPUs: RQ1 uses DeepSeek-V4-Pro~\cite{xu2026deepseek} as the judge backbone; RQ2 distills grouped models onto Qwen3.5-9B~\cite{qwen35blog} via Swift~\cite{swift}; and RQ3--RQ4 optimize Qwen3.6-35B-A3B~\cite{qwen36_35b_a3b} with Verl~\cite{verl}, vLLM~\cite{vllm}, and Megatron~\cite{megatron} (details in Appendix~\ref{app:impl_details}).

\subsection{Effectiveness of LLM Judge Interfaces (RQ1)}
To validate whether different LLM judge interfaces faithfully implement signal-specific decision boundaries, we compare ATLAS Judge against three alternatives under the same judge backbone: Direct Judge, Static Rubric, and Curated Rubric.
Table~\ref{tab:reward_judge_main} summarizes the held-out F1 of the LLM judge interfaces across the 41 LLM-based signals (full signal-level results in Appendix Tables~\ref{tab:reward_judge_full_a}--\ref{tab:reward_judge_full_b}).

\textbf{ATLAS Judge achieves the highest F1 across all reported groups, reaching 0.954 overall} (vs.\ 0.877 for Curated Rubric, 0.826 for Static Rubric, and 0.800 for Direct Judge).
The improvements are most pronounced on \textit{Timeliness} (0.961 vs.\ 0.691 for Direct Judge), \textit{Reliability} (0.972 vs.\ 0.811), and \textit{Intent \& Planning} (0.947 vs.\ 0.768), where accurate judgment requires cross-checking multi-step tool outputs, temporal validity, or query-rewrite constraints rather than inspecting the surface response alone.
This pattern shows that binding evaluation to admissible evidence, explicit exemptions, and stepwise review is critical: Direct Judge is competitive on local checks, whereas calibrated interfaces are essential when judgments depend on multi-step traces or exception-sensitive business rules.

\subsection{Diagnostic Model Distillation (RQ2)}
We next examine whether efficient diagnostic models can reproduce the decision behavior of selected LLM-based diagnostic signals.
Following Section~\ref{sec:model_distillation}, we distill grouped Qwen3.5-9B models for 17 high-frequency LLM-based signals across \textit{Relevance}, \textit{Factuality}, \textit{Reliability}, and \textit{Norms \& Compliance} (Table~\ref{tab:rm_distill_main}).
\input{tables/table_rm_distill_main}

Table~\ref{tab:rm_distill_main} shows that \textbf{the distilled 9B diagnostic models reach 94.6\% average F1, retaining most of the DeepSeek-V4-Pro reference performance} (96.4\%) while outperforming the untuned Qwen3.5-9B backbone (81.0\%, +13.6 points) and larger open baselines including Qwen3.6-27B (90.3\%) and Qwen3.6-35B-A3B (84.9\%).
The main insight is not compression alone, but calibrated decision-boundary transfer: once the student is trained against a well-calibrated LLM judge interface, a smaller model can outperform larger general-purpose models whose decision boundaries are not aligned to the target signal semantics.
The largest gains appear on business-grounded factuality and reliability signals where evidence binding and boundary control cannot be recovered from raw model capacity alone, clarifying the complementary roles in ATLAS: ATLAS Judge serves as the high-capacity reference evaluator, while distilled models provide low-latency, low-cost evaluation at scale.

\FloatBarrier
\Needspace{14\baselineskip}
\subsection{Human Audit of Agent Iteration (RQ3)}
\input{tables/table_human_audit_main}
We next evaluate whether the ATLAS-guided iteration of the Xiaotuan agent (\S\ref{sec:agent_iteration}; Appendix~\ref{app:impl_details}) improves service quality on unseen requests.
Prior to online rollout, pre-deployment diagnostic replay confirms consistent score improvements across within-turn execution dimensions and the across-turn user-wise layer, with the largest gain in \textit{Tool \& Skill Execution} (Appendix~\ref{app:offline_policy_eval}).
To evaluate response quality independently of ATLAS's own judges, we conduct a paired, blinded human audit over $1,000$ production queries sampled across local-services scenarios, request complexity levels, and multi-turn interaction types.
Sharing the same Qwen3.6-35B-A3B backbone, tool interfaces, and agent harness, both the deployed baseline and the ATLAS-guided agent concurrently generate complete tool-use trajectories and final responses for each query and its dialogue context.
Blinded domain annotators (with independent dual review and expert adjudication on disagreements) inspect each response against the user request and retrieved supply evidence, assigning binary $0/1$ judgments on three criteria: \textbf{response--supply relevance} ($1$ if the recommended supply satisfies the primary intent and key constraints), \textbf{P0 hallucination} ($1$ if the core conclusion or key factual claims contain unsupported or fabricated facts), and \textbf{ID hallucination} ($1$ if any core recommendation ID is fabricated, mismatched, or untraceable to tool evidence).

Table~\ref{tab:human_audit_main} summarizes the human-audit results.
Compared with the deployed baseline, the ATLAS-guided agent increases response--supply relevance by \textbf{+6.2 pp} while reducing P0 hallucination rate and ID hallucination rate by \textbf{$-$6.9 pp} and \textbf{$-$2.0 pp}, respectively.
These blinded human judgments confirm that targeting the execution and grounding deficiencies identified by ATLAS yields more relevant and factually reliable recommendations.

\FloatBarrier
\Needspace{20\baselineskip}
\input{tables/table_online_eval_main}
\subsection{Online Evaluation (RQ4)}
To assess real-world service impact at scale, we evaluate the ATLAS-guided agent against the previously deployed production baseline in a 14-day online A/B test on live Meituan Xiaotuan traffic from August 14 to August 27, with cumulative enrollment of nearly two million users per arm.
Using user-level hash bucketing with equal traffic splits, control and treatment arms are evaluated synchronously over the two-week window under two-sided testing.

Table~\ref{tab:online_eval_main} reports the relative changes ($\Delta$), $p$-values, and 95\% confidence intervals (CI) across six core engagement and transaction metrics.
The ATLAS-guided agent achieves statistically significant improvements ($p<0.0001$) across user engagement, supply exposure, and paid order volume: AI answer read-through rate increases by \textbf{+2.84\%}, effective query volume (QV) by \textbf{+0.89\%}, merchant-card exposure QV by \textbf{+4.23\%}, and paid order count by \textbf{+6.13\%}, while 5-second early session exit rate decreases by \textbf{$-$1.29\%} (with paid GTV showing a positive trend of $+4.51\%$).
These gains reflect a consistent service progression, as improved intent alignment and tool grounding reduce early abandonment, surface more actionable merchant cards, and align with higher downstream order volume.
Together with the human audit in RQ3, these results confirm that ATLAS-guided iteration translates capability diagnosis into measurable gains in user experience and downstream transactions, and the improved agent is now deployed to all production traffic.

%% file: tables/table_reward_judge_main.tex
\begin{table*}[!t]
    \centering
    \caption{Comparison of LLM judge interfaces across diagnostic signal groups in RQ1 (average F1 on $\mathcal{D}^{\mathrm{test}}_m$).}
    \vspace{-6pt}
    \label{tab:reward_judge_main}
    \setlength{\tabcolsep}{3.8pt}
    \resizebox{\linewidth}{!}{
    \begin{tabular}{lcccccccc}
        \toprule
        \textbf{Method} & \textbf{Relevance} & \textbf{Factuality} & \textbf{Timeliness} & \textbf{Reliability} & \textbf{Intent \& Planning} & \textbf{Norms \& Compliance} & \textbf{User-wise} & \textbf{Overall} \\
        \midrule
        Direct Judge & 0.831 & 0.842 & 0.691 & 0.811 & 0.768 & 0.813 & 0.784 & 0.800 \\
        Static Rubric & 0.843 & 0.853 & 0.849 & 0.775 & 0.811 & 0.876 & 0.761 & 0.826 \\
        Curated Rubric & 0.885 & 0.905 & 0.838 & 0.808 & 0.843 & 0.932 & 0.880 & 0.877 \\
        \rowcolor[gray]{0.92}\textbf{ATLAS Judge} & \textbf{0.939} & \textbf{0.956} & \textbf{0.961} & \textbf{0.972} & \textbf{0.947} & \textbf{0.984} & \textbf{0.953} & \textbf{0.954} \\
        \bottomrule
    \end{tabular}}
    \vspace{-6pt}
\end{table*}

%% file: tables/table_rm_distill_main.tex
\begin{table*}[!t]
    \centering
    \caption{Diagnostic model distillation results for the 17 selected signals in RQ2 (F1 in \% on $\mathcal{D}^{\mathrm{test}}_m$; $\Delta$ vs.\ untuned Qwen3.5-9B).}
    \vspace{-6pt}
    \label{tab:rm_distill_main}
    \scriptsize
    \setlength{\tabcolsep}{4pt}
    \renewcommand{\arraystretch}{1.08}
    \resizebox{0.98\linewidth}{!}{%
    \begin{tabular}{llcccccc}
        \toprule
        \textbf{Signal group} & \textbf{Diagnostic signal} & \textbf{DS-V4-Pro} & \textbf{Qwen3.5-9B} & \textbf{Qwen3.6-27B} & \textbf{Qwen3.6-35B-A3B} & \cellcolor[gray]{0.94}\textbf{Ours} & \cellcolor[gray]{0.96}\textbf{$\Delta$} \\
        \midrule
        \multirow{6}{*}{Relevance}
        & Content Relevance & 92.8 & 80.5 & 88.9 & 88.1 & \cellcolor[gray]{0.94}92.3 & \cellcolor[gray]{0.96}+11.8 \\
        & Supply Diversity & 94.3 & 83.1 & 93.4 & 91.9 & \cellcolor[gray]{0.94}93.5 & \cellcolor[gray]{0.96}+10.4 \\
        & Supply Relevance & 94.8 & 78.6 & 91.0 & 88.4 & \cellcolor[gray]{0.94}91.3 & \cellcolor[gray]{0.96}+12.7 \\
        & Opening Relevance & 97.8 & 72.6 & 92.6 & 89.5 & \cellcolor[gray]{0.94}94.3 & \cellcolor[gray]{0.96}+21.7 \\
        & Primary Rec. Match & 96.4 & 69.5 & 88.2 & 82.9 & \cellcolor[gray]{0.94}91.9 & \cellcolor[gray]{0.96}+22.4 \\
        & \textit{Avg.} & 95.2 & 76.9 & 90.8 & 88.2 & \cellcolor[gray]{0.94}92.7 & \cellcolor[gray]{0.96}+15.8 \\
        \midrule
        \multirow{7}{*}{Factuality}
        & Supply Existence & 93.8 & 89.6 & 84.4 & 87.3 & \cellcolor[gray]{0.94}94.5 & \cellcolor[gray]{0.96}+4.9 \\
        & Attribute Consistency & 91.0 & 67.5 & 76.2 & 80.2 & \cellcolor[gray]{0.94}89.0 & \cellcolor[gray]{0.96}+21.5 \\
        & Supply Attribution & 92.5 & 68.5 & 81.6 & 84.4 & \cellcolor[gray]{0.94}92.4 & \cellcolor[gray]{0.96}+23.9 \\
        & Failure Transparency & 99.8 & 94.1 & 97.1 & 83.3 & \cellcolor[gray]{0.94}94.4 & \cellcolor[gray]{0.96}+0.3 \\
        & POI Traceability & 99.7 & 99.2 & 86.3 & 90.0 & \cellcolor[gray]{0.94}98.6 & \cellcolor[gray]{0.96}-0.6 \\
        & Location Constraint & 98.6 & 69.4 & 79.2 & 68.7 & \cellcolor[gray]{0.94}98.6 & \cellcolor[gray]{0.96}+29.2 \\
        & \textit{Avg.} & 95.9 & 81.4 & 84.1 & 82.3 & \cellcolor[gray]{0.94}94.6 & \cellcolor[gray]{0.96}+13.2 \\
        \midrule
        \multirow{3}{*}{Reliability}
        & Logical Contradiction & 96.7 & 82.6 & 95.5 & 90.2 & \cellcolor[gray]{0.94}96.1 & \cellcolor[gray]{0.96}+13.5 \\
        & Semantic Repetition & 97.7 & 73.3 & 95.6 & 80.9 & \cellcolor[gray]{0.94}94.8 & \cellcolor[gray]{0.96}+21.5 \\
        & \textit{Avg.} & 97.2 & 78.0 & 95.5 & 85.6 & \cellcolor[gray]{0.94}95.5 & \cellcolor[gray]{0.96}+17.5 \\
        \midrule
        \multirow{5}{*}{Norms \& Compliance}
        & Format Validity & 98.9 & 86.2 & 97.9 & 97.1 & \cellcolor[gray]{0.94}96.1 & \cellcolor[gray]{0.96}+9.9 \\
        & Security \& Privacy & 99.6 & 95.7 & 99.2 & 95.1 & \cellcolor[gray]{0.94}98.9 & \cellcolor[gray]{0.96}+3.2 \\
        & System Safety & 98.1 & 89.9 & 98.8 & 71.1 & \cellcolor[gray]{0.94}98.4 & \cellcolor[gray]{0.96}+8.5 \\
        & Compliance & 97.0 & 77.5 & 89.0 & 74.4 & \cellcolor[gray]{0.94}93.5 & \cellcolor[gray]{0.96}+16.0 \\
        & \textit{Avg.} & 98.4 & 87.3 & 96.2 & 84.4 & \cellcolor[gray]{0.94}96.7 & \cellcolor[gray]{0.96}+9.4 \\
        \midrule
        \textbf{Overall} & \textbf{Avg.} & \textbf{96.4} & \textbf{81.0} & \textbf{90.3} & \textbf{84.9} & \cellcolor[gray]{0.92}\textbf{94.6} & \cellcolor[gray]{0.94}\textbf{+13.6} \\
        \bottomrule
    \end{tabular}%
    }
    \vspace{-6pt}
\end{table*}

%% file: tables/table_human_audit_main.tex
\begin{wraptable}{r}{0.50\textwidth}
    \centering
    \captionsetup{font=footnotesize}
    \caption{Blinded human audit on 1,000 paired production queries in RQ3, reporting changes ($\Delta$, in percentage points, pp) of the ATLAS-guided agent over the deployed baseline.}
    \label{tab:human_audit_main}
    \footnotesize
    \setlength{\tabcolsep}{3pt}
    \renewcommand{\arraystretch}{1.12}
    \begin{tabularx}{\linewidth}{@{}>{\raggedright\arraybackslash}Xcc@{}}
        \toprule
        \textbf{Human-audit metric} & \textbf{Direction} & \textbf{$\Delta$} \\
        \midrule
        Response--supply relevance & $\uparrow$ & \textbf{+6.2 pp} \\
        P0 hallucination rate & $\downarrow$ & \textbf{$-$6.9 pp} \\
        ID hallucination rate & $\downarrow$ & \textbf{$-$2.0 pp} \\
        \bottomrule
    \end{tabularx}

\end{wraptable}

%% file: tables/table_online_eval_main.tex
\begin{table}[H]
    \centering
    \caption{14-day online A/B test results on live Meituan Xiaotuan traffic in RQ4 (nearly two million users per arm; relative change $\Delta$, two-sided $p$-value, and 95\% CI).}
    \label{tab:online_eval_main}

    \small
    \setlength{\tabcolsep}{8pt}
    \renewcommand{\arraystretch}{1.15}
    \begin{tabular}{@{}lcccc@{}}
        \toprule
        \textbf{Online metric} & \textbf{Dir.} & \textbf{$\Delta$} & \textbf{$p$-value} & \textbf{CI} \\
        \midrule
        AI answer read-through rate & $\uparrow$ & \textbf{+2.84\%} & $<0.0001$ & $[+2.27\%, +3.38\%]$ \\
        Effective QV & $\uparrow$ & \textbf{+0.89\%} & $<0.0001$ & $[+0.56\%, +1.23\%]$ \\
        Merchant-card exposure QV & $\uparrow$ & \textbf{+4.23\%} & $<0.0001$ & $[+3.81\%, +4.66\%]$ \\
        Paid order count & $\uparrow$ & \textbf{+6.13\%} & $<0.0001$ & $[+2.89\%, +8.67\%]$ \\
        Paid GTV & $\uparrow$ & \textbf{+4.51\%} & $0.1355$ & $[-1.41\%, +10.43\%]$ \\
        5-second session exit rate & $\downarrow$ & \textbf{$-$1.29\%} & $<0.0001$ & $[-1.45\%, -1.19\%]$ \\
        \bottomrule
    \end{tabular}

\end{table}

%% file: sections/5-conclusion.tex
\section{Conclusion}
In this work, we introduced \textbf{ATLAS}, a dual-horizon diagnostic evaluation framework for industrial tool-use agents.
Rather than treating agent evaluation as outcome reporting alone, ATLAS organizes interpretable behavioral evidence to support capability analysis and targeted iteration in production settings.
Through complementary within-turn and across-turn evaluation, it jointly examines trajectory-wise evidence within a complete agent lifecycle and user-wise evidence across continued interaction.
Fine-grained diagnostic signals make these views executable through explicit evidence scopes and decision boundaries, while dynamically calibrated LLM judges and grouped compact diagnostic models support evaluation at scale.
The resulting diagnostic profiles help development teams attribute failures across execution stages, prioritize improvement targets, and reassess subsequent interventions.
Applied in Meituan Xiaotuan, a million-scale DAU local-services assistant, ATLAS achieves high diagnostic fidelity across calibrated LLM judges and distilled models.
Independent human auditing and a 14-day online A/B test with nearly two million enrolled users per arm further verify significant gains in recommendation quality, user engagement, and paid orders, with the improved agent deployed to all production traffic across web-scale tool-use services.

%% file: sections/appendix.tex
\section{Xiaotuan Platform and Agent Workflow}
\label{app:xiaotuan_platform}
Meituan Xiaotuan is an AI assistant for local services that supports user decision-making, service inquiry, and transactions across food delivery, in-store dining, hotel and travel, instant retail, and healthcare.
Unlike traditional keyword search that returns raw item lists for manual filtering, Xiaotuan handles open-ended or multi-constraint requests (e.g., location, budget, ambiance, party size, and dietary needs) through an iterative agentic workflow:
(i) \emph{intent understanding and clarification} to parse explicit and implicit constraints or ask follow-up questions;
(ii) \emph{tool and skill execution} across diverse production interfaces (such as location search, merchant and deal retrieval, menu and inventory queries, review mining, and domain-specific skills);
(iii) \emph{observation and reflection} to inspect tool outputs and reformulate queries when candidates are insufficient or expired; and
(iv) \emph{response synthesis} to deliver grounded answers and transactable merchant or deal cards across multi-turn sessions.

\FloatBarrier

\section{Additional Judge-Interface Results}

\subsection{Judge Interface Configurations}
\label{app:judge_interface_configurations}
All RQ1 variants use the same DeepSeek-V4-Pro backbone and are evaluated on the same held-out test splits ($\mathcal{D}^{\mathrm{test}}_m$).
Starting from the calibrated \textbf{ATLAS Judge} specification (refined on $\mathcal{D}^{\mathrm{cal}}_m$ with explicit evidence binding, exemption boundaries, stepwise review, and constrained output formatting), the three baselines are constructed via controlled component removal or simplification: \textbf{Direct Judge} retains only the task definition, input description, and output schema; \textbf{Static Rubric} augments the basic specification with an automatically generated fixed checklist; and \textbf{Curated Rubric} compresses signal-specific criteria and key exemptions into a compact rubric with a simplified review procedure.

\subsection{Fine-Grained LLM Judge-Interface Results}
Tables~\ref{tab:reward_judge_full_a} and~\ref{tab:reward_judge_full_b} report the signal-level results underlying the RQ1 main table.
All variants share the same backbone and are evaluated on the corresponding held-out test splits ($\mathcal{D}^{\mathrm{test}}_m$).
Together, the tables compare Direct Judge, Static Rubric, Curated Rubric, and ATLAS Judge on all LLM-based diagnostic signals.

\input{tables/table_reward_judge_full}

\section{Online Product Metric Definitions}
\label{app:metrics}
For the 14-day online A/B evaluation in RQ4, we report the following automatic product metrics measured from live traffic:
\begin{itemize}[leftmargin=*, itemsep=1pt, topsep=2pt]
    \item \textbf{AI answer read-through rate:} the fraction of sessions in which the AI response is read through to the end or reaches the platform-defined completion criterion.
    \item \textbf{Effective QV:} the volume of valid AI-assistant visits that satisfy the platform's effectiveness criterion.
    \item \textbf{Merchant-card exposure QV:} the volume of visits in which structured merchant or supply cards are exposed to the user.
    \item \textbf{Paid order count:} the total number of paid orders attributed to AI-assistant interactions.
    \item \textbf{Paid GTV:} the total paid gross transaction volume attributed to AI-assistant interactions.
    \item \textbf{5-second session exit rate:} the fraction of sessions that terminate within 5 seconds after entering the AI-assistant interaction.
\end{itemize}

\section{Additional Implementation Details}
\label{app:impl_details}
All experiments are conducted on NVIDIA H20 GPUs. For LLM judge-interface validation, all compared interfaces use the same DeepSeek-V4-Pro backbone~\cite{xu2026deepseek}; the calibration pipeline further relies on Claude Opus-4.6, GPT-5.4, and Gemini 3.1-Pro to construct high-confidence supervision, following Section~\ref{sec:signal_calibration}.
For diagnostic model distillation (\S\ref{sec:model_distillation}), we train grouped Qwen3.5-9B~\cite{qwen35blog} student models with Swift~\cite{swift} on 8 GPUs using teacher supervision from DeepSeek-V4-Pro, DeepSeek-V4-Flash~\cite{xu2026deepseek}, and GLM-5.1~\cite{zeng2026glm}. We use AdamW with batch size 128, learning rate $1\times10^{-5}$, maximum output length 2048, and maximum sequence length 16384.

For RL, we optimize Qwen3.6-35B-A3B~\cite{qwen36_35b_a3b} with the GRPO-based training pipeline in Section~\ref{sec:agent_iteration}. Training runs on 32 GPUs with Verl~\cite{verl}, using Megatron~\cite{megatron} for distributed training and vLLM~\cite{vllm} for rollout acceleration. The batch size is 128, the rollout group size is 16, the sampling temperature is 1.0, the maximum output length is 8192, and the maximum total sequence length is 32768. We use AdamW with learning rate $1\times10^{-6}$.
For the gated training return in Equation~(\ref{eq:training_reward}), signal roles and per-signal weights $\lambda_m$ are configured according to the diagnostic profile and business priorities as follows:
\begin{itemize}[leftmargin=*]
    \item \textbf{Guardrail gate ($G_{\mathrm{norm}}(x,\tau)$):} All signals in \emph{Norms \& Compliance} serve as a hard binary gate; failure on any guardrail signal sets the total return $R(x,\tau)$ to $0$.
    \item \textbf{Primary response quality ($\lambda_m=0.2$ per signal):} All \emph{Relevance} and \emph{Factuality} signals in \emph{Response Generation}.
    \item \textbf{Supporting execution, timeliness, and cross-turn quality ($\lambda_m=0.1$ per signal):} The remaining signals in \emph{Response Generation}, the \emph{Factuality} signals in \emph{Tool \& Skill Execution} (\emph{POI Traceability}, \emph{Location Constraint}, and \emph{Tool-Call Hallucination}), the \emph{Timeliness} signals in \emph{Thinking \& Reflection} (\emph{Outdated Supply Detection}) and \emph{Tool \& Skill Execution} (\emph{Supply Timeliness}), and the across-turn \emph{User-wise} signals.
\end{itemize}
For each input $x$, GRPO samples $G=16$ trajectories $\{\tau_i\}_{i=1}^{G}$ and computes the group-normalized advantage $\hat{A}_i = (R(x,\tau_i)-\mu_R)/(\sigma_R+\epsilon)$, where $\mu_R$ and $\sigma_R$ denote the mean and standard deviation of returns across the $G$ trajectories. Training over the $10,000$ requests in $\mathcal{D}_{\text{RL}}$ updates parameters once per $128$ requests (approximately $80$ steps per pass, reaching a plateau around step $60$).

\Needspace{18\baselineskip}
\section{Offline Diagnostic Replay Breakdown}
\label{app:offline_policy_eval}
\begin{wrapfigure}{r}{0.50\textwidth}
    \centering
    \captionsetup{font=footnotesize}
    \includegraphics[width=\linewidth]{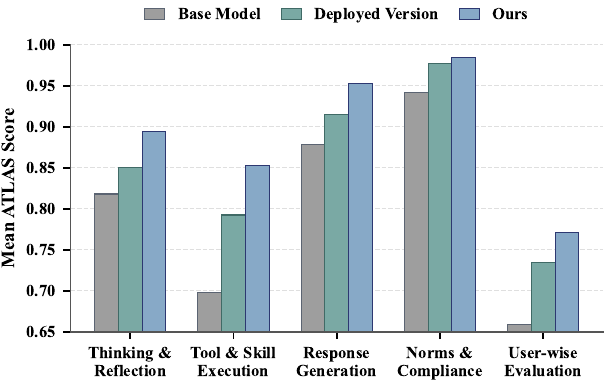}
    \caption{Pre-deployment offline replay evaluation under ATLAS across within-turn execution dimensions, the Norms \& Compliance guardrail, and the across-turn user-wise layer.}
    \label{fig:offline_eval_blocks}
\end{wrapfigure}

Prior to online rollout, we conduct a pre-deployment diagnostic check on $2,000$ conversation-disjoint production requests ($\mathcal{D}_{\text{Policy-Eval}}$), where each tested agent executes tool calls through production interfaces against live business supply.
Figure~\ref{fig:offline_eval_blocks} reports group-level mean ATLAS scores across the three within-turn execution dimensions, the \textit{Norms \& Compliance} guardrail, and the user-wise evaluation layer. Ours achieves the highest score in every reported group: it improves both trajectory-wise execution and response generation, maintains the high \textit{Norms \& Compliance} guardrail score, and improves across-turn user-wise performance. The largest improvement appears in \textit{Tool \& Skill Execution}, indicating that the gains extend beyond surface response phrasing to intermediate tool grounding and execution behavior, in agreement with the human audit (RQ3) and online A/B test (RQ4).

\FloatBarrier
\section{ATLAS Diagnostic Signal Inventory}
Tables~\ref{tab:reward_catalog_generation}--\ref{tab:reward_catalog_userwise} list all within-turn trajectory-wise signals (across the three execution dimensions and the Norms \& Compliance guardrail) and across-turn user-wise signals, including each signal's concern, type, and definition.

\begingroup
\makeatletter
\setlength{\@fptop}{0pt}
\setlength{\@fpsep}{14pt}
\setlength{\@fpbot}{0pt plus 1fil}
\setlength{\@dblfptop}{0pt}
\setlength{\@dblfpsep}{14pt}
\setlength{\@dblfpbot}{0pt plus 1fil}
\makeatother
\input{tables/table_reward_catalog_generation}
\input{tables/table_reward_catalog_tool}
\input{tables/table_reward_catalog_thinking}
\input{tables/table_reward_catalog_norms}
\input{tables/table_reward_catalog_userwise}

\clearpage
\endgroup

%% file: tables/table_reward_judge_full.tex
\begin{table}[!htbp]
    \centering
    \begin{minipage}[t]{0.49\linewidth}
    \centering
    \begin{minipage}[t][44pt][t]{\linewidth}
    \captionsetup{font=footnotesize}
    \caption{Fine-grained F1 comparison of LLM judge interfaces (Part I: Relevance, Factuality, Timeliness, and Reliability).}
    \label{tab:reward_judge_full_a}
    \end{minipage}\par
    \scriptsize
    \setlength{\tabcolsep}{2pt}
    \renewcommand{\arraystretch}{1.08}
    \begin{tabular*}{\linewidth}{@{\extracolsep{\fill}}>{\raggedright\arraybackslash}m{3.5cm}cccc@{}}
        \toprule
        \textbf{Signal} & \textbf{Direct} & \textbf{Static} & \textbf{Curated} & \textbf{ATLAS} \\
        \midrule
        \multicolumn{5}{l}{\textbf{Relevance}} \\
        Opening Relevance & 0.940 & 0.957 & 0.965 & \textbf{0.978} \\
        Primary Recommendation Match & 0.654 & 0.780 & 0.873 & \textbf{0.964} \\
        Content Relevance & 0.876 & 0.860 & 0.902 & \textbf{0.928} \\
        Supply Relevance & 0.926 & 0.904 & 0.934 & \textbf{0.948} \\
        Supply Diversity & 0.942 & 0.929 & 0.921 & \textbf{0.943} \\
        Evidence Insufficiency & 0.786 & 0.899 & 0.907 & \textbf{0.935} \\
        Supply Information Insufficiency & 0.661 & 0.725 & 0.833 & \textbf{0.918} \\
        Irrelevant Supply & 0.861 & 0.690 & 0.745 & \textbf{0.901} \\
        \rowcolor[gray]{0.94}\textit{Avg.} & 0.831 & 0.843 & 0.885 & \textbf{0.939} \\
        \midrule
        \multicolumn{5}{l}{\textbf{Factuality}} \\
        Reflection Hallucination & 0.843 & 0.758 & 0.905 & \textbf{0.947} \\
        Rejection Hallucination & 0.933 & 0.873 & 0.954 & \textbf{0.989} \\
        Tool-Call Hallucination & 0.757 & 0.774 & 0.777 & \textbf{0.910} \\
        Supply Existence & 0.878 & 0.917 & 0.890 & \textbf{0.938} \\
        Attribute Consistency & 0.851 & 0.893 & 0.872 & \textbf{0.910} \\
        Supply Attribution & 0.885 & 0.817 & 0.907 & \textbf{0.925} \\
        Failure Transparency & 0.739 & 0.905 & 0.948 & \textbf{0.998} \\
        POI Traceability & 0.911 & 0.952 & 0.955 & \textbf{0.997} \\
        Location Constraint & 0.784 & 0.791 & 0.936 & \textbf{0.986} \\
        \rowcolor[gray]{0.94}\textit{Avg.} & 0.842 & 0.853 & 0.905 & \textbf{0.956} \\
        \midrule
        \multicolumn{5}{l}{\textbf{Timeliness}} \\
        Outdated Supply Detection & 0.777 & 0.942 & 0.957 & \textbf{0.970} \\
        Response Timeliness & 0.843 & 0.927 & 0.866 & \textbf{0.970} \\
        Supply Timeliness & 0.454 & 0.678 & 0.690 & \textbf{0.944} \\
        \rowcolor[gray]{0.94}\textit{Avg.} & 0.691 & 0.849 & 0.838 & \textbf{0.961} \\
        \midrule
        \multicolumn{5}{l}{\textbf{Reliability}} \\
        Logical Contradiction & 0.844 & 0.920 & 0.938 & \textbf{0.967} \\
        Semantic Repetition & 0.778 & 0.630 & 0.677 & \textbf{0.977} \\
        \rowcolor[gray]{0.94}\textit{Avg.} & 0.811 & 0.775 & 0.808 & \textbf{0.972} \\
        \bottomrule
    \end{tabular*}
    \end{minipage}\hfill
    \begin{minipage}[t]{0.49\linewidth}
    \centering
    \begin{minipage}[t][44pt][t]{\linewidth}
    \captionsetup{font=footnotesize}
    \caption{Fine-grained F1 comparison of LLM judge interfaces (Part II: Intent \& Planning, Norms \& Compliance, and User-wise Evaluation).}
    \label{tab:reward_judge_full_b}
    \end{minipage}\par
    \scriptsize
    \setlength{\tabcolsep}{2pt}
    \renewcommand{\arraystretch}{1.08}
    \begin{tabular*}{\linewidth}{@{\extracolsep{\fill}}>{\raggedright\arraybackslash}m{3.5cm}cccc@{}}
        \toprule
        \textbf{Signal} & \textbf{Direct} & \textbf{Static} & \textbf{Curated} & \textbf{ATLAS} \\
        \midrule
        \multicolumn{5}{l}{\textbf{Intent \& Planning}} \\
        Query Calibration & 0.673 & 0.615 & 0.786 & \textbf{0.911} \\
        Reflection Gain & 0.682 & 0.749 & 0.679 & \textbf{0.916} \\
        Argument Validity & 0.515 & 0.778 & 0.806 & \textbf{0.931} \\
        Rewrite Validity & 0.818 & 0.750 & 0.800 & \textbf{0.932} \\
        Constraint Coverage & 0.899 & 0.925 & 0.936 & \textbf{0.962} \\
        Search Intent Understanding & 0.852 & 0.868 & 0.908 & \textbf{0.972} \\
        Skill Behavior & 0.764 & 0.857 & 0.841 & \textbf{0.962} \\
        Skill Selection & 0.809 & 0.804 & 0.902 & \textbf{0.970} \\
        Tool-Call Order & 0.901 & 0.949 & 0.930 & \textbf{0.966} \\
        \rowcolor[gray]{0.94}\textit{Avg.} & 0.768 & 0.811 & 0.843 & \textbf{0.947} \\
        \midrule
        \multicolumn{5}{l}{\textbf{Norms \& Compliance}} \\
        Compliance & 0.798 & 0.739 & 0.931 & \textbf{0.970} \\
        Security \& Privacy & 0.895 & 0.874 & 0.888 & \textbf{0.996} \\
        Format Validity & 0.709 & 0.967 & 0.975 & \textbf{0.989} \\
        System Safety & 0.848 & 0.922 & 0.933 & \textbf{0.981} \\
        \rowcolor[gray]{0.94}\textit{Avg.} & 0.813 & 0.876 & 0.932 & \textbf{0.984} \\
        \midrule
        \multicolumn{5}{l}{\textbf{User-wise Evaluation}} \\
        Cross-Turn Reference & 0.905 & 0.898 & 0.938 & \textbf{0.970} \\
        Reasoning-Intent Alignment & 0.913 & 0.784 & 0.910 & \textbf{0.939} \\
        Tool-Intent Alignment & 0.901 & 0.856 & 0.917 & \textbf{0.941} \\
        Previous-Round Correction & 0.378 & 0.482 & 0.643 & \textbf{0.938} \\
        Cross-Round Contradiction & 0.911 & 0.884 & 0.946 & \textbf{0.970} \\
        Cross-Round Repetition & 0.695 & 0.663 & 0.926 & \textbf{0.958} \\
        \rowcolor[gray]{0.94}\textit{Avg.} & 0.784 & 0.761 & 0.880 & \textbf{0.953} \\
        \bottomrule
    \end{tabular*}
    \end{minipage}
\end{table}

%% file: tables/table_reward_catalog_generation.tex
\begin{table*}[htbp]
    \centering
    \caption{Within-turn evaluation: trajectory-wise diagnostic signals for Response Generation.}
    \label{tab:reward_catalog_generation}
    \scriptsize
    \setlength{\tabcolsep}{4pt}
    \renewcommand{\arraystretch}{1.08}
    \resizebox{\linewidth}{!}{%
    \begin{tabular}{llll}
        \toprule
        \textbf{Concern} & \textbf{Diagnostic signal} & \textbf{Type} & \textbf{Definition} \\
        \midrule
        \multirow{5}{*}{Relevance}
        & Opening Relevance & LLM & Checks whether the opening paragraph directly addresses the user request. \\
        & Primary Rec. Match & LLM & Checks whether the primary recommended supply item is the most suitable one. \\
        & Content Relevance & LLM & Evaluates whether the response content stays on topic and answers the intended need. \\
        & Supply Relevance & LLM & Evaluates whether the referenced merchants, POIs, or supply items are relevant and well ranked. \\
        & Supply Diversity & LLM & Checks whether the selected supply items provide sufficiently diverse useful options. \\
        \cmidrule(lr){1-4}
        \multirow{5}{*}{Factuality}
        & Supply Existence & LLM & Detects hallucinated supply existence such as non-existent merchants or POIs. \\
        & Attribute Consistency & LLM & Checks whether response claims remain consistent with supply attributes and evidence. \\
        & Supply Attribution & LLM & Detects entity or merchant misattribution in the final response. \\
        & Failure Transparency & LLM & Checks whether the response honestly reports tool failure instead of fabricating results. \\
        & Rejection Hallucination & LLM & Detects unsupported factual claims inside fallback or refusal responses. \\
        \cmidrule(lr){1-4}
        Timeliness
        & Response Timeliness & LLM & Detects whether the response cites stale or temporally invalid supply information. \\
        \cmidrule(lr){1-4}
        \multirow{3}{*}{Reliability}
        & Logical Contradiction & LLM & Detects internal logical contradiction in the final delivered response. \\
        & Semantic Repetition & LLM & Detects severe semantic repetition or rambling in the final response. \\
        & N-gram Repetition & Rule & Detects low-level repeated spans and degenerate repetitive generation. \\
        \bottomrule
    \end{tabular}%
    }
\end{table*}

%% file: tables/table_reward_catalog_tool.tex
\begin{table*}[htbp]
    \centering
    \caption{Within-turn evaluation: trajectory-wise diagnostic signals for Tool \& Skill Execution.}
    \label{tab:reward_catalog_tool}
    \scriptsize
    \setlength{\tabcolsep}{4pt}
    \renewcommand{\arraystretch}{1.08}
    \resizebox{\linewidth}{!}{%
    \begin{tabular}{llll}
        \toprule
        \textbf{Concern} & \textbf{Diagnostic signal} & \textbf{Type} & \textbf{Definition} \\
        \midrule
        \multirow{2}{*}{Relevance}
        & Irrelevant Supply & LLM & Detects retrieved supply that is completely irrelevant to the request. \\
        & Supply Info Insufficiency & LLM & Checks whether retrieved supply contains the information required to answer the request. \\
        \cmidrule(lr){1-4}
        \multirow{3}{*}{Factuality}
        & POI Traceability & LLM & Checks whether POI or merchant identifiers can be traced to valid tool evidence. \\
        & Location Constraint & LLM & Checks whether location-sensitive parameters remain faithful to evidence and user request. \\
        & Tool-Call Hallucination & LLM & Detects invented slots, unsupported arguments, or fabricated tool-call content. \\
        \cmidrule(lr){1-4}
        Timeliness
        & Supply Timeliness & LLM & Evaluates whether retrieved supply is temporally valid and up to date. \\
        \cmidrule(lr){1-4}
        \multirow{2}{*}{Reliability}
        & Tool-Call Schema & Rule & Checks whether the tool-call schema and parameter structure are legal. \\
        & Tool-Call Turn Count & Rule & Monitors whether the number of tool-call turns stays within a reasonable range. \\
        \cmidrule(lr){1-4}
        \multirow{6}{*}{\shortstack[c]{Intent \&\\Planning}}
        & Constraint Coverage & LLM & Evaluates whether search or retrieval arguments correctly cover the required objective constraints. \\
        & Rewrite Validity & LLM & Evaluates whether a rewritten search query preserves the original intent. \\
        & Argument Validity & LLM & Checks whether tool arguments are well formed and semantically valid. \\
        & Skill Selection & LLM & Evaluates whether the selected skill or route is appropriate for the task. \\
        & Skill Behavior & LLM & Evaluates whether the activated skill follows its expected behavioral protocol. \\
        & Tool-Call Order & LLM & Checks whether tool invocation order follows required workflow constraints. \\
        \bottomrule
    \end{tabular}%
    }
\end{table*}

%% file: tables/table_reward_catalog_thinking.tex
\begin{table*}[htbp]
    \centering
    \caption{Within-turn evaluation: trajectory-wise diagnostic signals for Thinking \& Reflection.}
    \label{tab:reward_catalog_thinking}
    \scriptsize
    \setlength{\tabcolsep}{4pt}
    \renewcommand{\arraystretch}{1.08}
    \resizebox{\linewidth}{!}{%
    \begin{tabular}{llll}
        \toprule
        \textbf{Concern} & \textbf{Diagnostic signal} & \textbf{Type} & \textbf{Definition} \\
        \midrule
        Relevance
        & Evidence Insufficiency & LLM & Checks whether the reflection step correctly identifies insufficient evidence. \\
        \cmidrule(lr){1-4}
        Factuality
        & Reflection Hallucination & LLM & Detects whether the reflection step invents unsupported observations. \\
        \cmidrule(lr){1-4}
        Timeliness
        & Outdated Supply Detection & LLM & Checks whether reflection correctly identifies stale or outdated supply. \\
        \cmidrule(lr){1-4}
        Reliability
        & CoT Token Count & Rule & Monitors reasoning-trace length as a coarse operational reliability signal. \\
        \cmidrule(lr){1-4}
        \multirow{3}{*}{\shortstack[c]{Intent \&\\Planning}}
        & Search Intent Understanding & LLM & Evaluates whether the model correctly understands the user intent at the first turn. \\
        & Query Calibration & LLM & Checks whether the model corrects noisy or malformed user intent when needed. \\
        & Reflection Gain & LLM & Measures whether reflection or react reasoning adds useful information. \\
        \bottomrule
    \end{tabular}%
    }
\end{table*}

%% file: tables/table_reward_catalog_norms.tex
\begin{table*}[htbp]
    \centering
    \caption{Within-turn evaluation: trajectory-wise diagnostic signals for the Norms \& Compliance guardrail.}
    \label{tab:reward_catalog_norms}
    \scriptsize
    \setlength{\tabcolsep}{4pt}
    \renewcommand{\arraystretch}{1.08}
    \resizebox{\linewidth}{!}{%
    \begin{tabular}{lll}
        \toprule
        \textbf{Diagnostic signal} & \textbf{Type} & \textbf{Definition} \\
        \midrule
        Compliance & LLM & Checks whether the model properly refuses impossible, invalid, or non-completable requests. \\
        Security \& Privacy & LLM & Detects privacy leakage, unsafe disclosure, or other security-sensitive violations. \\
        System Safety & LLM & Checks whether the final response satisfies system-level safety requirements. \\
        Format Validity & LLM & Checks non-trivial structural and formatting validity beyond simple rules. \\
        Structural Gate & Rule & Enforces mandatory structural validity and required formatting constraints. \\
        Presentation Integrity & Rule & Enforces secondary presentation and formatting constraints. \\
        \bottomrule
    \end{tabular}%
    }
\end{table*}

%% file: tables/table_reward_catalog_userwise.tex
\begin{table*}[htbp]
    \centering
    \caption{Across-turn evaluation: user-wise diagnostic signals.}
    \label{tab:reward_catalog_userwise}
    \scriptsize
    \setlength{\tabcolsep}{4pt}
    \renewcommand{\arraystretch}{1.08}
    \resizebox{\linewidth}{!}{%
    \begin{tabular}{lll}
        \toprule
        \textbf{Diagnostic signal} & \textbf{Type} & \textbf{Definition} \\
        \midrule
        Cross-Turn Reference & LLM & Evaluates whether the model resolves referential ambiguity using prior interaction context. \\
        Reasoning-Intent Alignment & LLM & Evaluates whether subsequent-lifecycle reasoning remains aligned with the user intent. \\
        Tool-Intent Alignment & LLM & Evaluates whether subsequent-lifecycle tool use remains aligned with the intended task. \\
        Previous-Round Correction & LLM & Checks whether the model corrects a previously misunderstood intent. \\
        Cross-Round Contradiction & LLM & Detects contradiction across dialogue turns or reasoning steps. \\
        Cross-Round Repetition & LLM & Detects severe repetition across turns. \\
        \bottomrule
    \end{tabular}%
    }
\end{table*}